\documentclass[sigconf]{acmart}

\usepackage{tabularx}
\usepackage{booktabs}
\acmConference[FIRE '24]{Forum for Information Retrieval Evaluation}{12th - 15th December 2024}{DA-IICT, Gandhinagar, India}

\setcopyright{acmlicensed}
\copyrightyear{2024}
\acmYear{2024}
\acmDOI{XXXXXXX.XXXXXXX}
\acmISBN{978-1-4503-XXXX-X/18/06}

\AtBeginDocument{%
  \providecommand\BibTeX{{%
    \normalfont B\kern-0.5em{\scshape i\kern-0.25em b}\kern-0.8em\TeX}}}

\begin{document}

\title{On Importance of Layer Pruning for Smaller BERT Models and Low Resource Languages}


\author{Mayur Shirke}
\email{shirkemayur49@gmail.com}
\affiliation{%
  \institution{Pune Institute of Computer Technology, Pune}
  \country{India}
}
\affiliation{%
  \institution{L3Cube Labs, Pune}
  \city{Pune}
  \state{Maharashtra}
  \country{India}
}

\author{Amey Shembade}
\email{ameyshembade2511@gmail.com}
\affiliation{%
  \institution{Pune Institute of Computer Technology, Pune}
  \country{India}
}
\affiliation{%
  \institution{L3Cube Labs, Pune}
  \city{Pune}
  \state{Maharashtra}
  \country{India}
}

\author{Madhushri Wagh}
\email{madhushriwagh15@gmail.com}
\affiliation{%
  \institution{Pune Institute of Computer Technology, Pune}
  \country{India}
}
\affiliation{%
  \institution{L3Cube Labs, Pune}
  \city{Pune}
  \state{Maharashtra}
  \country{India}
}

\author{Pavan Thorat}
\email{thoratpavan64@gmail.com}
\affiliation{%
  \institution{Pune Institute of Computer Technology, Pune}
  \country{India}
}
\affiliation{%
  \institution{L3Cube Labs, Pune}
  \city{Pune}
  \state{Maharashtra}
  \country{India}
}

\author{Raviraj Joshi}
\email{ravirajoshi@gmail.com}
\affiliation{%
  \institution{Indian Institute of Technology Madras, Chennai}
  \country{India}
}
\affiliation{%
  \institution{L3Cube Labs, Pune}
  \city{Pune}
  \state{Maharashtra}
  \country{India}
}







\begin{abstract}
  This study explores the effectiveness of layer pruning for developing more efficient BERT models tailored to specific downstream tasks in low-resource languages. Our primary objective is to evaluate whether pruned BERT models can maintain high performance while reducing model size and complexity. We experiment with several BERT variants, including MahaBERT-v2 and Google-Muril, applying different pruning strategies and comparing their performance to smaller, scratch-trained models like MahaBERT-Small and MahaBERT-Smaller. We fine-tune these models on Marathi datasets, specifically Short Headlines Classification (SHC), Long Paragraph Classification (LPC) and Long Document Classification (LDC), to assess their classification accuracy. Our findings demonstrate that pruned models, despite having fewer layers, achieve comparable performance to their fully-layered counterparts while consistently outperforming scratch-trained models of similar size. Notably, pruning layers from the middle of the model proves to be the most effective strategy, offering performance competitive with pruning from the top and bottom. However, there is no clear winner, as different pruning strategies perform better in different model and dataset combinations. Additionally, monolingual BERT models outperform multilingual ones in these experiments. This approach, which reduces computational demands, provides a faster and more efficient alternative to training smaller models from scratch, making advanced NLP models more accessible for low-resource languages without compromising classification accuracy.
\end{abstract}




\maketitle

\section{Introduction}
Recent advancements in language models have underscored the need for efficient and scalable solutions, particularly when applied to low-resource languages. While Transformer-based models like BERT \cite{devlin2018bertpretrain} (Bidirectional Encoder Representations from Transformers) have set new benchmarks in natural language processing (NLP), their large size and computational requirements pose significant challenges, especially for low-resource language \cite{chin2018layer} classification tasks.

BERT's architecture utilizes a multi-layer bidirectional Transformer encoder that has demonstrated remarkable performance across a variety of NLP tasks. However, its extensive parameterization and computational demands often hinder its deployment in resource-constrained environments. To address these challenges, model compression techniques such as layer pruning have gained traction \cite{ganesh2021compressing}. Layer pruning involves selectively removing parts of a neural network \cite{cheng2021survey} to reduce its size and computational burden while preserving performance.

\begin{figure}[htbp]
  \centering
  \includegraphics[width=\linewidth]{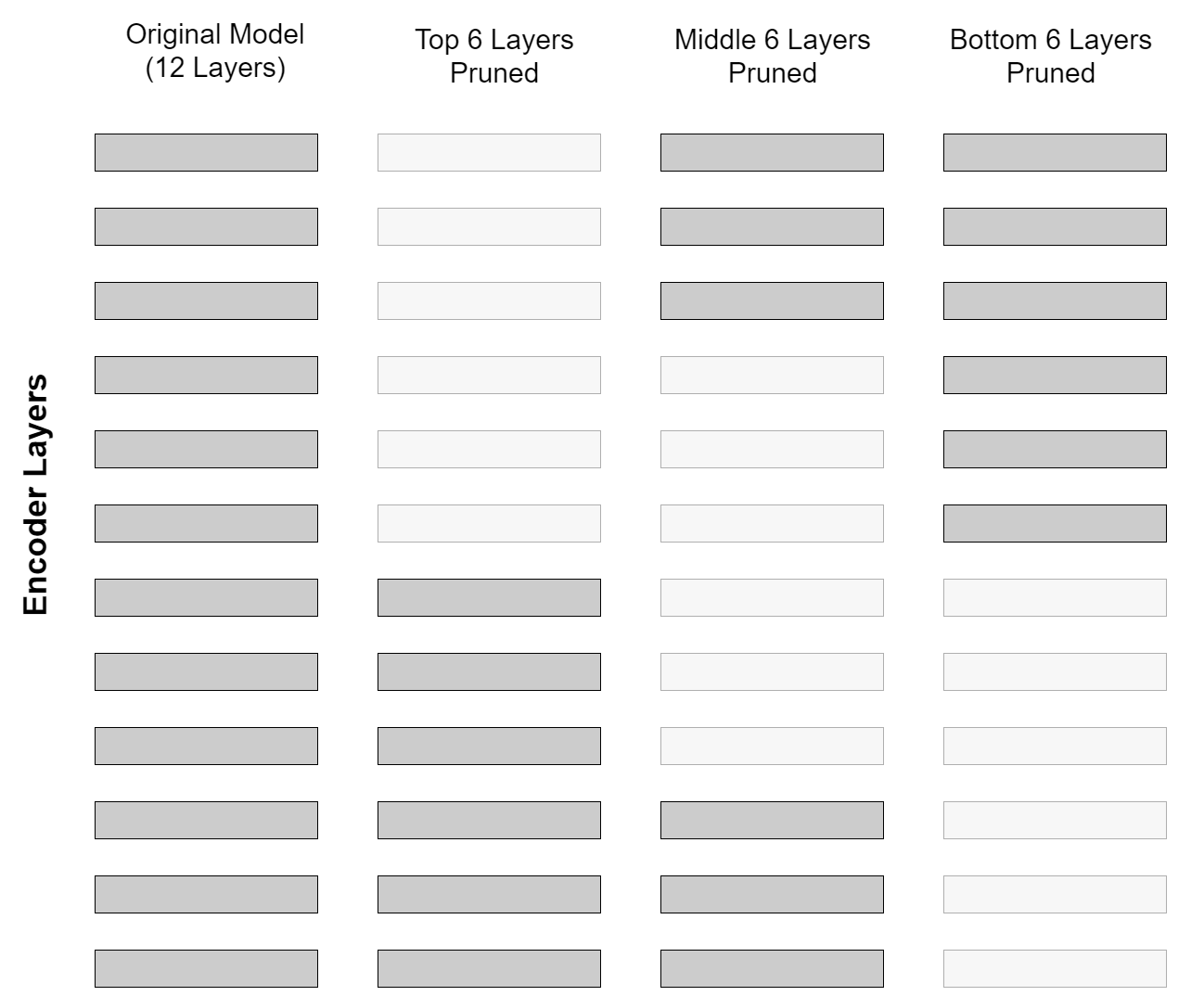}
  \caption{Layer pruning strategies for BERT models.}
  \Description{Illustration of different layer pruning strategies for BERT models, showing how layers are selectively removed to optimize the model.}
  \label{fig1}
\end{figure}

This study explores the impact of layer pruning \cite{michael2017} on smaller BERT models for text classification in the low-resource Marathi language. We evaluate various pruning strategies, such as top, middle, and bottom layer pruning \cite{hassan2022}, and find that different strategies perform better depending on the model and dataset combination, with no clear overall winner. Among these, pruning from the middle emerges as a promising choice, offering competitive scores compared to top and bottom pruning.

Using datasets from the L3Cube-IndicNews Corpus, we evaluate MahaBERT-v2 and Google-Muril models against their pruned variants and baseline models such as MahaBERT-Small and MahaBERT-Smaller. Our results show that middle-layer pruning offers the best balance, achieving 50\% to 80\% size reduction while maintaining competitive performance. These findings highlight the adaptability of pruning strategies for resource-constrained \cite{chin2018layer} settings. While prior work has focused on SBERT pruning for embedding tasks, this study provides new insights into BERT pruning for text classification. Our 2-layer and 6-layer pruned MahaBERT-v2 models consistently outperform similarly sized scratch-trained models like MahaBERT-Small and MahaBERT-Smaller.

These results suggest that layer pruning, combined with fine-tuning, is an efficient, cost-effective strategy for developing smaller, high-performing task-specific BERT models \cite{fan2019layerdrop}. This approach is particularly beneficial for low-resource languages, offering a more feasible alternative to the computationally expensive process of pre-training smaller BERT models from scratch.

The main contributions of the work are as follows:
\begin{itemize}
\item We propose an efficient method for creating smaller, task-specific BERT models for low-resource languages. By using layer pruning followed by fine-tuning for specific classification tasks, we demonstrate that this approach yields more efficient and high-performing models. This method is also significantly faster than pre-training smaller BERT models from scratch before performing task-specific fine-tuning.
\item We show that pruning layers from the middle of the model is an optimal strategy in most cases, with scores competitive to pruning from the top and bottom.
\item Different pruning strategies perform better depending on model and dataset combinations, indicating no clear winner across all scenarios.
\item Monolingual BERT models outperform multilingual BERT models under the experimental settings presented in this paper.
\end{itemize}

\section{Related Work}

Recent research on CNN pruning has explored various methods to reduce computational and storage costs while maintaining performance. Traditional approaches include weight pruning based on magnitude and gradient-based methods \cite{yeom2018pruning}. More recent developments involve pruning criteria derived from explainable AI techniques, such as Layer-wise Relevance Propagation (LRP), which evaluates the importance of network units by tracing their contribution to the final decision. This method offers a promising alternative to conventional pruning techniques, especially in scenarios where fine-tuning is limited or data is scarce.

Previous work on CNN pruning has primarily focused on methodologies such as pruning connections, channels, and filters to reduce model complexity. Techniques like network slimming and structured pruning aim to eliminate redundant parameters while maintaining accuracy. For instance, methods proposed by \cite{pasandi2018modeling} involve pruning less significant weights and fine-tuning the model to recover performance. These foundational techniques for reducing CNN model size and computational cost are directly relevant to our work.

Building on these traditional methods, a novel approach proposed in \cite{chen2019shallowing} focuses on feature representation-based layer-wise pruning. Unlike conventional weight-based pruning, this method identifies redundant parameters by analyzing learned features in convolutional layers. This layer-level pruning significantly reduces computational costs while maintaining or improving model performance across various datasets.

Another advanced method, Layer-Compensated Pruning (LcP), is introduced in \cite{chin2018layer} to optimize CNNs for resource-constrained environments, such as mobile and edge devices. This approach integrates layer scheduling and filter ranking into a global filter ranking strategy, optimized using meta-learning. Evaluated on VGG, ResNet, and MobileNetV2, LcP improves pruning efficiency and accuracy, narrowing the accuracy gap between pruned and original models while significantly reducing meta-learning time.

The work on deep neural network pruning extends to transformer-based models as well. Transformer models like BERT have achieved remarkable performance across various NLP tasks but are often resource-intensive due to their large number of parameters. To address this, research has focused on model compression techniques, including pruning, quantization, and knowledge distillation, which reduce memory consumption and computational costs without significant performance loss \cite{ganesh2021compressing}. These methods enable the creation of smaller, faster, and more efficient versions of models like BERT.

LayerDrop introduces a structured dropout technique that regularizes transformer models by dropping entire layers during training. This method allows for the extraction of sub-networks of varying depths from a single over-parameterized model without requiring fine-tuning. By pruning layers in this manner, LayerDrop reduces computational costs while maintaining strong performance across various NLP tasks, including machine translation and language modeling. This approach simplifies pruning, offering a more efficient alternative to traditional techniques like distillation \cite{fan2019layerdrop}.

In the domain of Indic NLP, \cite{aggarwal2022indicxnli} introduces the INDICXNLI dataset, which extends the English XNLI dataset to 11 Indic languages using high-quality machine translation. The work evaluates cross-lingual transfer using models such as MuRIL and XLM-RoBERTa, with MuRIL outperforming in most cases except for English-centric tasks. This study highlights the challenges faced by low-resource languages like Odia and suggests integrating IndicXNLI into IndicGLUE to broaden its evaluation scope. Training on both English and Indic languages improves performance, particularly for low-resource languages.

Finally, \cite{devlin2018bertpretrain} introduces BERT, a model that utilizes deep bidirectional Transformers. Key innovations include the Masked Language Model (MLM) and Next Sentence Prediction (NSP), which enhance contextual understanding and text pair representations. BERT achieves state-of-the-art results across 11 NLP tasks, outperforming previous models in low-resource scenarios and various benchmarks, demonstrating the significance of bidirectionality in language models.

\section{Methodologies}
In this study, we investigate the effectiveness of layer pruning in optimizing BERT models for low-resource text classification tasks. Our primary objective is to reduce model complexity while maintaining or improving performance, particularly for Marathi language classification datasets. We focus on three different datasets: Short Headlines Classification (SHC), Long Paragraph Classification (LPC), and Long Document Classification (LDC). These datasets vary in content length and structure, providing a robust testing ground for evaluating pruning strategies.

 The BERT models we experimented with include the Maha BERT-v2, Google-Muril, and smaller versions of Marathi BERT models. We evaluate multiple pruning strategies, including top-layer pruning, middle-layer pruning, and bottom-layer pruning, aiming to identify which sections of the model can be pruned without significantly impacting classification accuracy. The models are fine-tuned on the SHC, LPC, and LDC datasets to classify news articles or sub-articles into predefined categories.
 
 To ensure the reliability of our results, we report both validation and testing accuracies across different pruning configurations. We also compare pruned models against baseline models trained from scratch to highlight the advantages of pruning, especially when working with low-resource languages. 

\subsection{Models and Datasets}

\subsubsection{Models}
Below, we outline the various models utilized in our experiments, detailing their architectures and the pruning techniques implemented:

\begin{itemize}
    \item \textbf{MahaBERT-v2:} Maha BERT-v2, a specialized BERT variant fine-tuned for Marathi, consists of 12 transformer layers and is tailored for diverse Marathi text processing applications. To evaluate the balance between model size and performance, especially for low-resource classification tasks, we experimented with various pruning strategies. These included removing sets of 6 and 10 layers from the top, middle, and bottom of the model.\cite{joshi2022}
    
    \item \textbf{Google-Muril:}Google-Muril \cite{simran2021}, a multilingual BERT variant trained on diverse linguistic datasets including Marathi, is equipped with 12 transformer layers and supports a wide array of text processing tasks. To analyze its performance across multiple languages, we employed various pruning strategies, such as removing 6 and 10 layers from the top, middle, and bottom of the model.
    
    \item \textbf{Marathi BERT-Small:}Marathi BERT-Small is a lightweight version of the Marathi BERT model, comprising 6 transformer layers. This foundational BERT model is built from scratch using the L3Cube-MahaCorpus along with other publicly available Marathi monolingual datasets.\cite{joshi2022}
    
    \item \textbf{Marathi BERT-Smaller:} Marathi BERT-Smaller is an even more compact version of the Marathi BERT model, featuring only 2 transformer layers. It is independently trained from scratch using the L3Cube-MahaCorpus, supplemented by other publicly available Marathi monolingual datasets.\cite{joshi2022}
\end{itemize}

\subsubsection{Datasets}

We evaluated the performance of these models on three distinct datasets:

\begin{itemize}
    \begin{table}[htbp]
    \centering
    \caption{Category-wise distribution for SHC, LDC, and LPC datasets}
    \begin{tabular}{lccc}
    \toprule
    \textbf{Category} & \textbf{SHC Total} & \textbf{LDC Total} & \textbf{LPC Total} \\
    \midrule
    Train      & 22014  & 22014  & 22014 \\
    Test       & 2761   & 2761   & 2761  \\
    Validation & 2750   & 2750   & 2750  \\
    \midrule
    \textbf{Total} & \textbf{27525} & \textbf{27525} & \textbf{27525} \\
    \bottomrule
    \end{tabular}
    \end{table}

    \item \textbf{Short Headlines Classification (SHC):} The SHC dataset comprises news article headlines along with their respective categorical labels. Given the concise nature of the records, models are challenged to classify accurately with minimal context. This dataset is well-suited for testing model efficiency in handling short text classification tasks.\cite{joshi2024}
    
    \item \textbf{LPC (Long Paragraph Classification):} The LPC dataset consists of extended paragraphs taken from news articles, each paired with a categorical label. Unlike the SHC dataset, LPC offers more context, enabling models to leverage richer information for classification. It serves as a benchmark to assess model performance on medium-length texts where contextual details play a significant role in classification.\cite{joshi2024}
    
    \item \textbf{LDC (Long Document Classification):} The LDC dataset comprises complete news articles, each paired with a categorical label. These lengthy documents demand that models process substantial information while preserving classification accuracy. This dataset is especially valuable for evaluating the impact of pruning strategies on model performance in handling extensive texts that require a thorough understanding of the content.\cite{joshi2024}
\end{itemize}

\subsection{Experiments}

We conducted experiments on three datasets: Short Headlines Classification (SHC), Long Paragraph Classification (LPC), and Long Document Classification (LDC). Each dataset presents unique challenges due to varying text lengths and complexity, which allowed us to evaluate the impact of pruning on models across different types of text inputs. We applied pruning strategies to several BERT-based models: Maha BERT-v2, Google-Muril, MahaBERT-Smaller, and MahaBERT-Small. The un-pruned versions of the models were used as baselines for comparison.

\subsubsection{Pruning Strategies}

We explored several layer pruning strategies to reduce the complexity of the BERT models while retaining as much performance as possible. The pruning strategies were as follows:

\begin{itemize}
    \item \textbf{Top Layer Pruning:} This strategy involves removing the upper layers of the model, which handle high-level abstractions of the input text. We tested the effect of pruning by removing the top 6 and top 10 layers from the models. The goal was to observe how removing the most abstract representations affected the model's ability to classify Marathi text.
    \item \textbf{Middle Layer Pruning:} In this approach, we removed layers from the middle of the transformer architecture. Middle layers act as a transition between lower-level and higher-level representations. We pruned 6 and 10 layers from the middle of the models to assess whether the middle layers could be discarded without a significant loss in performance.
    \item \textbf{Bottom Layer Pruning:} Here, we focused on removing the bottom layers of the model, which capture lower-level linguistic features such as word meanings and grammatical structures. We pruned the bottom 6 and bottom 10 layers to determine how much foundational information could be eliminated while still achieving effective classification results.
\end{itemize}

\subsubsection{Evaluation Metrics}

The effectiveness of the pruning strategies was measured using the following metrics:

\begin{itemize}
   \item \textbf{Validation Accuracy:} Measured on the validation set during training to track the model's performance over time.
    \item \textbf{Testing Accuracy:} Evaluated on a separate test set to assess the final model performance after pruning and fine-tuning.
    \item \textbf{Model Size:} We monitored the reduction in the number of layers and parameters to analyze the trade-offs between model size and classification accuracy.
\end{itemize}

These metrics allowed us to evaluate how well the pruned models performed in terms of both accuracy and efficiency across the SHC, LPC, and LDC datasets.

\section{Results}

The results obtained from fine-tuning the models on our datasets are presented in Table \ref{tab_combined}. The testing accuracy for all models and pruning strategies is provided in the table, highlighting the performance across different datasets. For detailed validation accuracy, please refer to Tables 2, 3, and 4 in the appendix. These tables provide a comprehensive view of the models' performance during validation, complementing the testing accuracy results.

The results indicate that while there is no clear winner among the different pruning strategies, performance varies depending on the model and dataset combination. In many cases, pruning from the middle demonstrates competitive scores compared to top and bottom pruning, emerging as a promising choice for achieving a balance between efficiency and accuracy.

\begin{table}[htbp]
  \caption{Performance of Models on Marathi SHC, LPC, and LDC Datasets (Testing Accuracy Only, in Percentage)}
  \label{tab_combined}
  \centering
  \begin{tabularx}{\linewidth}{lXccc}
    \toprule
    \textbf{Model} & \textbf{Pruning Strategy} & \textbf{SHC} & \textbf{LPC} & \textbf{LDC} \\
    \midrule
    MahaBERT-v2 & Top 6 & \textbf{92.18} & 90.80 & 89.35 \\
    MahaBERT-v2 & Middle 6 & 90.33 & 90.55 & 89.90 \\
    MahaBERT-v2 & Bottom 6 & 90.47 & \textbf{91.05} & \textbf{90.04} \\
    \midrule
    MahaBERT-v2 & Top 10 & 89.20 & \textbf{92.00} & 88.52 \\
    MahaBERT-v2 & Middle 10 & 89.13 & 89.68 & \textbf{89.49} \\
    MahaBERT-v2 & Bottom 10 & \textbf{89.35} & 89.71 & 88.99 \\
    \midrule
    Google-Muril & Top 6 & 89.08 & 88.92 & 89.28 \\
    Google-Muril & Middle 6 & \textbf{90.69} & \textbf{90.37} & 88.88 \\
    Google-Muril & Bottom 6 & 87.62 & 89.70 & \textbf{90.11} \\
    \midrule
    Google-Muril & Top 10 & 88.22 & 88.66 & 88.55 \\
    Google-Muril & Middle 10 & 88.55 & \textbf{90.00} & 89.06 \\
    Google-Muril & Bottom 10 & \textbf{89.18} & 89.13 & 88.92 \\
    \midrule
    MahaBERT-Smaller & \textemdash & 88.11 & 90.62 & 84.39 \\
    MahaBERT-Small & \textemdash & 88.81 & 89.46 & 85.04 \\
    \midrule
    MahaBERT-v2 & \textemdash & 91.41 & 88.75 & 94.78 \\
    Google-Muril & \textemdash & 90.11 & 86.58 & 93.02 \\
    \bottomrule
  \end{tabularx}
\end{table}

\subsection{Marathi SHC Dataset}
According to Table \ref{tab_combined}, in the Marathi SHC dataset, MahaBERT-v2 demonstrates superior performance when utilizing the Top 6 pruning strategy, achieving a peak testing accuracy of 92.18 \%. This surpasses the results obtained through other pruning methods, with Middle 6 and Bottom 6 yielding 90.33\% and 90.47\%, respectively. Conversely, Google-Muril exhibits optimal performance with the Middle 6 pruning approach, reaching an accuracy of 90.69\%, while its Top 6 and Bottom 6 strategies produce lower accuracies of 89.08\% and 87.62\%, respectively. The more compact Marathi BERT models show moderate results, with MahaBERT-Small attaining 88.81\% and MahaBERT-Smaller reaching 88.11\%.

\subsection{Marathi LPC Dataset}
Examining the Marathi LPC dataset results in Table \ref{tab_combined}, we observe that MahaBERT-v2 demonstrates superior performance when employing the Top 10 pruning strategy, achieving a peak testing accuracy of 92.00\%. The Bottom 6 approach also yields impressive results with 91.05\%, closely followed by the Middle 6 strategy at 90.55\%. For Google-Muril, the Middle 6 pruning technique proves most effective, reaching an accuracy of 90.37\%, while the Bottom 6 and Top 6 methods produce lower scores of 89.70\% and 88.92\%, respectively. Among the more compact models, MahaBERT-Smaller edges out MahaBERT-Small with testing accuracies of 90.62\% and 89.46\%, respectively.

\subsection{Marathi LDC Dataset}
In the Marathi LDC dataset results (Table \ref{tab_combined}), MahaBERT-v2 using the Bottom 6 pruning approach yields the highest test accuracy at 90.04\%. The Middle 6 strategy follows closely with 89.90\%, while the Top 6 pruning method for MahaBERT-v2 shows a slightly lower performance at 89.35\%. For Google-Muril, the Bottom 6 strategy leads with a test accuracy of 90.11\%, outperforming both the Top 6 and Middle 6 approaches, which achieve 89.28\% and 88.88\%, respectively. Among the compact models, MahaBERT-Small demonstrates superior performance with an accuracy of 85.04\%, surpassing MahaBERT-Smaller, which records a lower test accuracy of 84.39\%.

Overall, the results highlight that different pruning strategies perform better for different model and dataset combinations, with no clear winner. Pruning from the middle proves to be a generally effective choice, often achieving scores competitive with top and bottom pruning strategies.

\section{Conclusion}

Our research investigates the effectiveness of layer pruning in enhancing the performance and efficiency of compact BERT models for classification tasks in low-resource languages. The experimental results indicate that no single pruning strategy consistently outperforms others across all models and datasets. Different pruning strategies excel in different model and dataset combinations, reflecting the complexity of optimizing BERT models. However, pruning from the middle layers generally provides a balanced trade-off between model size reduction and performance retention, often achieving scores competitive with top and bottom pruning strategies.

This technique offers a viable approach to optimizing BERT models by significantly reducing computational demands while maintaining competitive accuracy. These findings suggest that layer pruning, particularly in the middle layers, can be a practical strategy for improving BERT’s efficiency in real-world applications without necessitating retraining from scratch.

We have shown that pruned models yield significant reductions in computational overhead while achieving strong accuracy by using pruning algorithms on models like Google-Muril and MahaBERT-v2. These results highlight layer pruning's promise as a workable approach for implementing cutting-edge NLP tools in resource-constrained situations.

This study has been conducted exclusively on Marathi text. A future direction for this work is to investigate whether these findings hold across other low-resource languages, broadening the applicability and insights of this research.

\section{Acknowledgements}
We would like to express our sincere gratitude to L3Cube, Pune, for their invaluable mentorship and support throughout this work. Their guidance and expertise have been instrumental in shaping our research and achieving meaningful outcomes. Their encouragement and constructive feedback have been pivotal in refining our ideas and presenting this work effectively. Additionally, ChatGPT was used to generate the text for this paper.


\bibliographystyle{ACM-Reference-Format}
\bibliography{main.bib}
\clearpage

\appendix 
\section*{Appendix A: Model Performance Tables}

\begin{table}[htbp]
  \caption{Performance of Models on Marathi SHC Dataset (Accuracies in Percentage)}
  \label{tab1}
  \centering
  \resizebox{\linewidth}{!}{%
  \begin{tabular}{lccccc}
    \toprule
    \textbf{Model} & \textbf{Pruning Strategy} & \textbf{Validation Accuracy} & \textbf{Testing Accuracy} \\
    \midrule
    MahaBERT-v2 & Top 6 & \textbf{91.92} & \textbf{92.18} \\
    MahaBERT-v2 & Middle 6 & 90.95 & 90.33 \\
    MahaBERT-v2 & Bottom 6 & 91.05 & 90.47 \\
    \midrule
    MahaBERT-v2 & Top 10 & 88.81 & 89.20 \\
    MahaBERT-v2 & Middle 10 & 88.92 & 89.13 \\
    MahaBERT-v2 & Bottom 10 & \textbf{89.64} & \textbf{89.35} \\
    \midrule
    Google-Muril & Top 6 & 88.51 & 89.08 \\
    Google-Muril & Middle 6 & \textbf{90.00} & \textbf{90.69} \\
    Google-Muril & Bottom 6 & 88.30 & 87.62 \\
    \midrule
    Google-Muril & Top 10 & 88.16 & 88.22 \\
    Google-Muril & Middle 10 & \textbf{89.49} & 88.55 \\
    Google-Muril & Bottom 10 & 88.81 & \textbf{89.18} \\
    \midrule
    Marathi BERT Smaller & \textemdash & 88.41 & 88.11 \\
    Marathi BERT Small & \textemdash & 88.95 & 88.81 \\
    \bottomrule
  \end{tabular}%
  }
\end{table}
\begin{table}[htbp]
  \caption{Performance of Models on Marathi LPC Dataset (Accuracies in Percentage)}
  \label{tab2}
  \centering
  \resizebox{\linewidth}{!}{%
  \begin{tabular}{lccccc}
    \toprule
    \textbf{Model} & \textbf{Pruning Strategy} & \textbf{Validation Accuracy} & \textbf{Testing Accuracy} \\
    \midrule
    MahaBERT-v2 & Top 6 & 92.18 & 90.80 \\
    MahaBERT-v2 & Middle 6 & 92.33 & 90.55 \\
    MahaBERT-v2 & Bottom 6 & \textbf{92.36} & \textbf{91.05} \\
    \midrule
    MahaBERT-v2 & Top 10 & 90.37 & \textbf{92.00} \\
    MahaBERT-v2 & Middle 10 & 91.31 & 89.68 \\
    MahaBERT-v2 & Bottom 10 & \textbf{91.45} & 89.71 \\
    \midrule
    Google-Muril & Top 6 & 90.80 & 88.92 \\
    Google-Muril & Middle 6 & \textbf{92.00} & \textbf{90.37} \\
    Google-Muril & Bottom 6 & 88.95 & 89.70 \\
    \midrule
    Google-Muril & Top 10 & 91.13 & 88.66 \\
    Google-Muril & Middle 10 & \textbf{91.20} & \textbf{90.00} \\
    Google-Muril & Bottom 10 & 90.40 & 89.13 \\
    \midrule
    Marathi BERT Smaller & \textemdash & 89.03 & 90.62 \\
    Marathi BERT Small & \textemdash & 91.16 & 89.46 \\
    \bottomrule
  \end{tabular}%
  }
\end{table}

\begin{table}[htbp]
  \caption{Performance of Models on Marathi LDC Dataset (Accuracies in Percentage)}
  \label{tab3}
  \centering
  \resizebox{\linewidth}{!}{%
  \begin{tabular}{lccccc}
    \toprule
    \textbf{Model} & \textbf{Pruning Strategy} & \textbf{Validation Accuracy} & \textbf{Testing Accuracy} \\
    \midrule
    MahaBERT-v2 & Top 6 & 89.53 & 89.35 \\
    MahaBERT-v2 & Middle 6 & 90.51 & 89.90 \\
    MahaBERT-v2 & Bottom 6 & \textbf{90.65} & \textbf{90.04} \\
    \midrule
    MahaBERT-v2 & Top 10 & 89.24 & 88.52 \\
    MahaBERT-v2 & Middle 10 & 89.38 & \textbf{89.49} \\
    MahaBERT-v2 & Bottom 10 & \textbf{89.60} & 88.99 \\
    \midrule
    Google-Muril & Top 6 & 89.09 & 89.28 \\
    Google-Muril & Middle 6 & \textbf{89.82} & 88.88 \\
    Google-Muril & Bottom 6 & 89.71 & \textbf{90.11} \\
    \midrule
    Google-Muril & Top 10 & \textbf{89.67} & 88.55 \\
    Google-Muril & Middle 10 & 89.13 & \textbf{89.06} \\
    Google-Muril & Bottom 10 & 89.64 & 88.92 \\
    \midrule
    Marathi BERT Smaller & \textemdash & 84.44 & 84.39 \\
    Marathi BERT Small & \textemdash & 84.73 & 85.04 \\
    \bottomrule
  \end{tabular}%
  }
\end{table}

\end{document}